\documentclass{article}

\usepackage{microtype}
\usepackage{graphicx}
\usepackage{booktabs} 
\usepackage{multirow} 
\usepackage{hyperref}
\usepackage[table,xcdraw]{xcolor}

\usepackage[accepted]{icml2024}

\usepackage{amsmath}
\usepackage{amssymb}
\usepackage{mathtools}
\usepackage{amsthm}
\usepackage{enumitem}
\usepackage{minitoc}

\usepackage{tocloft}

\usepackage[capitalize,noabbrev]{cleveref}

\theoremstyle{plain}

\theoremstyle{definition}

\theoremstyle{remark}

\usepackage[textsize=tiny]{todonotes}

\icmltitlerunning{Non-confusing Generation of Customized Concepts in Diffusion Models}

\begin{document}

\twocolumn[
\icmltitle{Non-confusing Generation of Customized Concepts in Diffusion Models}


\icmlsetsymbol{equal}{*}

\begin{icmlauthorlist}
\icmlauthor{Wang Lin}{zju,equal}
\icmlauthor{Jingyuan Chen}{zju,cor}
\icmlauthor{Jiaxin Shi}{hcc,equal}
\icmlauthor{Yichen Zhu}{zju}
\icmlauthor{Chen Liang}{ts}
\icmlauthor{Junzhong Miao}{hit}
\icmlauthor{Tao Jin}{zju}
\icmlauthor{Zhou Zhao}{zju}
\icmlauthor{Fei Wu}{zju}
\icmlauthor{Shuicheng Yan}{sky}
\icmlauthor{Hanwang Zhang}{sky,ntu}
\end{icmlauthorlist}

\icmlaffiliation{zju}{Zhejiang University}
\icmlaffiliation{sky}{Skywork AI, Singapore}
\icmlaffiliation{hcc}{Huawei Cloud Computing}
\icmlaffiliation{ntu}{Nanyang Technological University.}
\icmlaffiliation{ts}{Tsinghua University}
\icmlaffiliation{hit}{Harbin Institute of Technology}
\icmlaffiliation{cor}{Corresponding Author}

\icmlcorrespondingauthor{Correspondence to: Jingyuan Chen}{jingyuanchen@zju.edu.cn}

\icmlkeywords{Machine Learning, ICML}

\vskip 0.3in
]



\printAffiliationsAndNotice{\icmlEqualContribution} 

\begin{abstract}
We tackle the common challenge of inter-concept visual confusion in compositional concept generation using text-guided diffusion models (TGDMs). It becomes even more pronounced in the generation of customized concepts, due to the scarcity of user-provided concept visual examples. By revisiting the two major stages leading to the success of TGDMs---1) contrastive image-language pre-training (CLIP) for text encoder that encodes visual semantics, and 2) training TGDM that decodes the textual embeddings into pixels---we point that existing customized generation methods only focus on fine-tuning the second stage while overlooking the first one. To this end, we propose a simple yet effective solution called CLIF: contrastive image-language fine-tuning. Specifically, given a few samples of customized concepts, we obtain non-confusing textual embeddings of a concept by fine-tuning CLIP via contrasting a concept and the over-segmented visual regions of other concepts. 
Experimental results demonstrate the effectiveness of CLIF in preventing the confusion of multi-customized concept generation. Project page: \href{https://clif-official.github.io/clif/}{https://clif-official.github.io/clif}.
\end{abstract}

\section{Introduction}
\begin{figure*}
  \centering
  \includegraphics[scale=0.48]{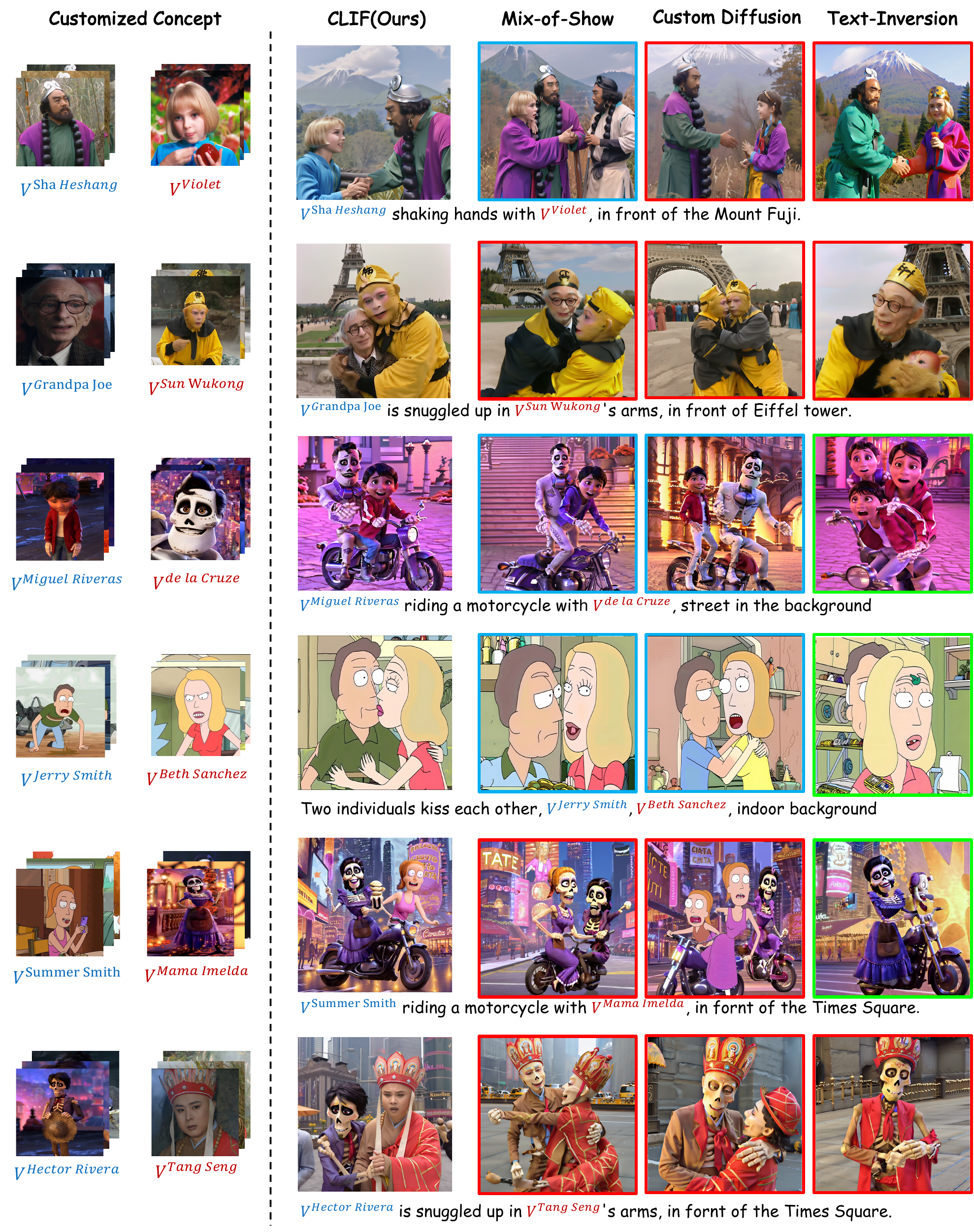}
\caption{\textbf{Visualization of multi-concept customization for challenging cases.} When the concepts to be customized belong to categories with high semantic similarity (all belonging to ``humans'' superclass), or when there is large regional overlap (\textit{e.g.}, the second and third rows) or combinations across styles (\textit{e.g.}, 2D combined with 3D characters in the fifth and sixth rows), 
the baseline methods suffer from, identity loss (red border), attribute leaking (blue border), or concept missing (green border), which are effectively circumvented by CLIF.}
\label{main_1}
\vspace{-1em}
\end{figure*}

We are interested in customizing a text-guided diffusion model (TGDM), \textit{e.g.}, Stable Diffusion~\cite{rombach2022high}, to generate compositions of user-provided concepts. For example, as shown in Figure~\ref{main_1}, given a few images of \textit{Hector Rivera} and \textit{Tang Seng}, we can generate imaginary compositions by the prompt ``\textit{Hector Rivera} snuggled up in \textit{Tang Seng}''. Existing customized generation methods are based on fine-tuning a pre-trained TGDM, where the tunable parameters include the textual embeddings of the new concept names~\cite{gal2022image,voynov2023p+} and/or LoRAs~\cite{hu2021lora} on the generation backbone~\cite{ruiz2023dreambooth,kumari2023multi}. In this way, the fine-tuned TGDM is expected to memorize the visual concepts and 
generalize them to unseen compositions. However, an ever-elusive challenge of the generalization is the inter-concept confusion shown in Figure~\ref{main_1}. This visual defect is even more severe when the interaction is spatially cluttered, such as ``snuggling'' and ``riding motorcycle''. 

\begin{figure}
  \centering
  \includegraphics[scale=0.3]{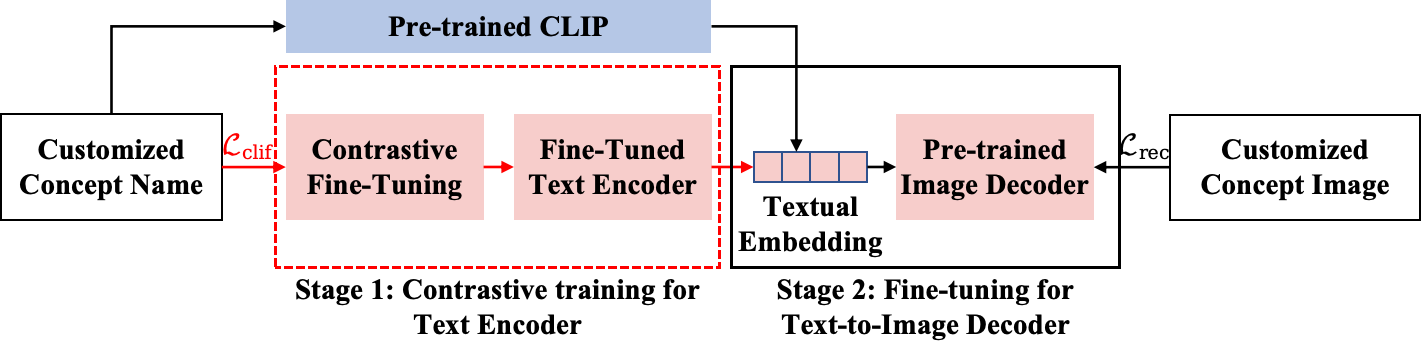}
  \vspace{-1.em}
\caption{The black line and box denote the prevailing pipeline of customized generation methods. Our contribution is to contrast the textual embeddings of customized concepts in the Text Encoder stage, which is shown in the red line and dashed box.}
\label{fig:1}
\vspace{-1em}
\end{figure}

Recent findings of visualizing the role of the image-text cross-attention in pre-trained TGDM~\cite{tewel2023key,patashnik2023localizing} show that the textual embeddings (V-values) control ``what to draw'' and the cross-attention map (Q-K softmax) tells ``where to draw''.
Inspired by this, we believe that the cause of the confusion is mainly due to the confusing textual embeddings of concepts. To see this, we revisit the two stages in training TGDM (Figure~\ref{fig:1} and Section~\ref{preliminary}): 
\begin{itemize}[noitemsep,nolistsep,leftmargin=*]
    \item \textbf{Text Encoder}: we obtain the text encoder from contrastive image-language pre-training (CLIP) on large-scale image-text pairs~\cite{radford2021learning, schuhmann2021laion}. In this way, the token embedding of a concept token carries its visual features. 
    \item \textbf{Text-to-Image Decoder}: the textual embedding is decoded into pixels by the cross-attention between textual and visual embeddings in the U-net decoder. In Section~\ref{related work}, we discuss that almost all the customized generation methods focus on this stage.
\end{itemize}

\begin{figure}
  \centering
  \includegraphics[scale=0.26]{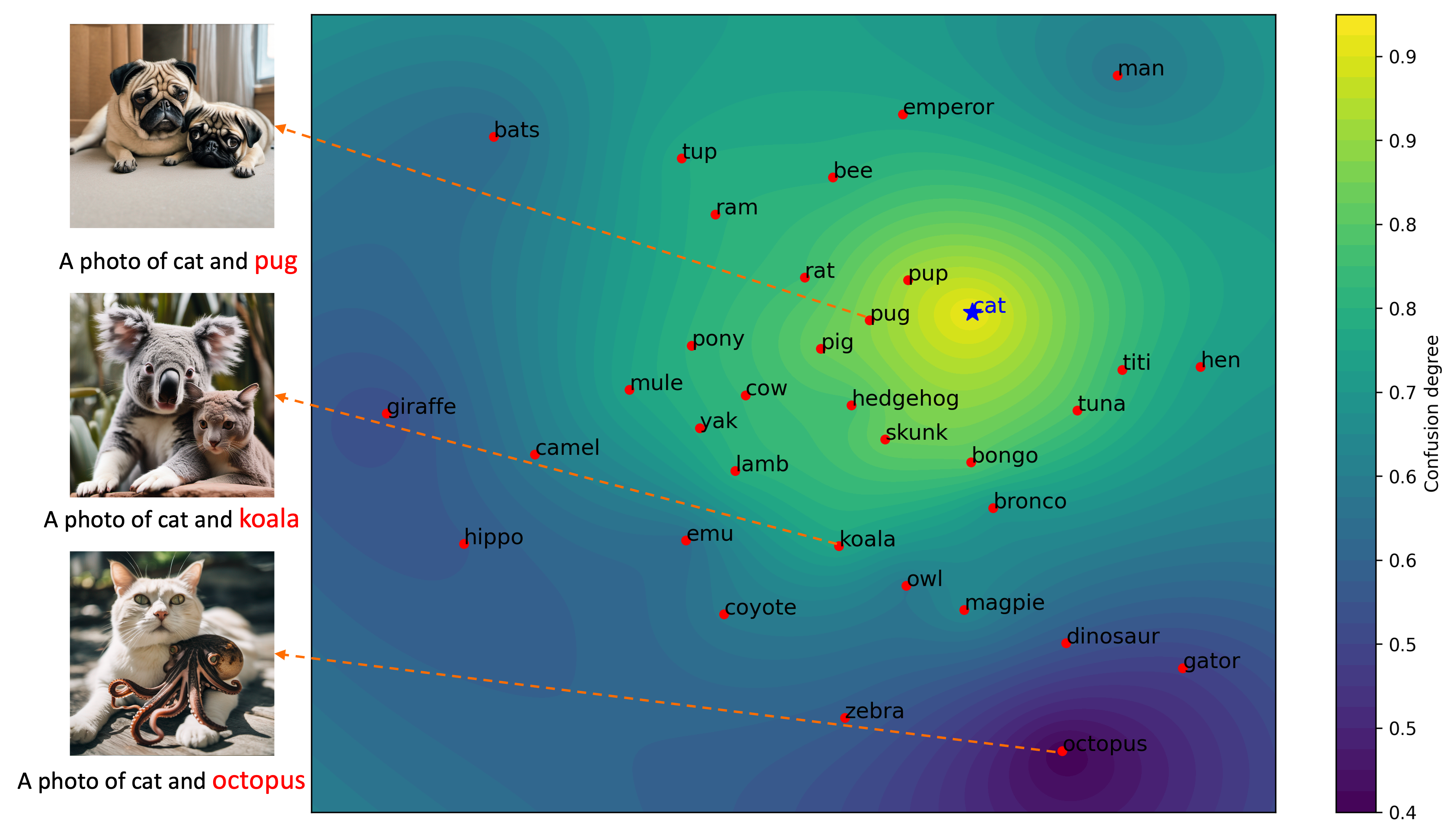}
\vspace{-1em}
\caption{Visualization of confusion in embedding space with ``cat" as an anchor point, see Appendix for details. It shows an evident correlation between confusion and embedding distance.}
\label{confusion map}
\vspace{-2em}
\end{figure}

We can show the confusion of common concepts in the existing vocabulary of Stable Diffusion by measuring the confusion degree of each concept. We use a sentence of two concepts as the prompt for generation (\textit{e.g.}, ``a cat and a dog''), and calculate the probability of their presence in the generated image using an object detector.
As illustrated in Figure~\ref{confusion map}, when the two concept token embeddings are far away (\textit{e.g.}, ``octopus'' and ``cat''), the composition is rarely confused; when they are close, confusion is common.

Can we use existing methods~\cite{chefer2023attend,li2023divide,huang2023composer,zhang2023adding,mou2023t2i} of de-confusing the above TGDM-known concepts to mitigate the confusion of TGDM-unknown, customized concepts? Unfortunately, the answer is no. This is because those methods rely on the assumption that the visual features and textual embeddings are well-aligned---but it does not hold for few-shot customized concepts. Then, what about the de-confusing methods especially designed for fine-tuning customized concepts~\cite{gu2023mix, po2023orthogonal}? Still no, because fine-tuning the second stage should not contrast the textual embeddings of different concepts too sharply to prevent overfitting, \textit{e.g.}, making the pre-trained TGDM lose the original ability of text control.

We propose a simple yet effective method called \textbf{CLIF}: Contrastive Language-Image \emph{Fine-tuning}, to tackle the confusion challenge directly in the first stage by contrasting the textual embeddings of customized concepts (Figure~\ref{fig:1}).  
We first present an over-segmented concept dataset that augments the visual examples of customized concepts into a large number of language-image contrastive data (Section~\ref{4.1}). Thus, by applying contrast fine-tuning of the text encoder on the augmented data, we can fundamentally eliminate the confusion in the concept token embeddings (Section~\ref{4.2}). Then, in the second stage, we reconstruct the images of concepts to fine-tune both the text embeddings with the text encoder frozen and the Unet of TGDM (Section~\ref{4.3}).

In Section~\ref{5}, to demonstrate the non-confusing effectiveness of CLIF, we jointly customize 18 user-provided characters and compare CLIF with prior SOTA methods as shown in Figure~\ref{main_1}. Different from prior methods~\cite{gu2023mix}, these 18 characters are more fair and proper for evaluation, because they are from less popular movies, which is rare for a pre-trained TGDM. We conduct extensive ablations to analyze how each CLIF's breakdowns mitigate confusion in multi-concept generation.

\section{Related Work}
\label{related work}
\noindent\textbf{Concept Customization}.
The goal of customized generation is to implant the user-provided visual examples of concepts into a pre-trained TGDM to generate various renditions of the concepts vividly guided by text prompts. Existing works can be categorized into three types based on the fine-tuned modules of the text-guided diffusion model (image decoder): 
1) Text embedding~\cite{gal2022image,voynov2023p+,yuan2023inserting,alaluf2023neural}: this type fine-tunes the text embeddings of customized concepts to align with input images while freezing the diffusion model. 
2) Decoder (\textit{e.g.}, U-net)~\cite{ruiz2023dreambooth}: this type explicitly binds the concept with rare words by fine-tuning the entire diffusion model. Additionally, ~\cite{ryu2023low} adopts a low-rank adapter (LoRA) ~\cite{hu2021lora} for concept tuning, which is lightweight and can achieve comparable fidelity to full-weight tuning.
3) Joint methods~\cite{kumari2023multi} that fine-tune the above two. However, these three modules are only in the image decoder stage of TGDM. In contrast, our CLIF recalls the neglected first stage and fine-tunes both of the stages, thereby improving the ability to customize and compose more diverse concepts.

\noindent\textbf{Confusion in Multi-Concept Customization.}
Some of the above works~\cite{kumari2023multi,gu2023mix,liu2023cones} 
focus on injecting multiple concepts into TGDM, facing the challenge of inter-concept confusion. 
To tackle the challenge, current approaches can be categorized into the following 3 types:
1) Generative semantic nursing~\cite{chefer2023attend, li2023divide, hertz2022prompt}: 
this type optimizes or edits the cross-attention maps in the generative process during inference time. 
Although for common concepts these methods can correct minor errors in attention maps and enable more accurate multi-concept synthesis, they are not salvageable for the confused attention maps generated by TGDM-unknown customized concepts.
2) Spatial control~\cite{huang2023composer,zhang2023adding,mou2023t2i}: this type directly integrates spatial layout as a pixel-level specification. However, obtaining additional spatial conditions is costly and proves ineffective in complex interactions involving large overlapping areas.
3) Token embedding~\cite{balaji2022ediffi, liu2022compositional, feng2022training}: this type aims to improve the prompt alignment on the text side by combining T5 and CLIP text encoders or utilizes language parsers to associate attributes solely with the corresponding concepts. In contrast, our approach does not require additional components, and we directly fine-tune the concept token embeddings to alleviate the confusion and achieve better prompt alignment.

\section{Preliminary}
\label{preliminary}

\textbf{CLIP as Text Encoder.}
For a concept with name $c$ (might be associated with some text prompts $p$ such as ``a photo of $c$''), the Text Encoder transforms it into a textual embedding $\text{V}^*$ that carries the concept's visual features as:
\begin{equation}
\centering
\begin{aligned}
\text{V}^* &:= \texttt{Text-Encoder}(p), \\
\end{aligned}
\end{equation}
where $:=$ denotes that $\text{V}^{*}$ is the token embedding selected from the position of $c$ in $p$.
Existing methods employ CLIP to train Text Encoder by contrastive pre-training on a large-scale dataset of text-image pairs.
Specifically, for each text-image pair $(p, q)$ in a batch, the goal of CLIP is to maximize the similarity between the text and the corresponding image while minimizing the similarity with another non-matching image. The training loss can be formulated as:
\begin{equation}
\centering
\begin{aligned}
\mathcal{L}_{clip}=-\log \frac{\exp(\text{s}(\texttt{Text-Encoder}(p), f(q)))}{\sum_{n^-} \exp(\text{s}(\texttt{Text-Encoder}(p), f(q^-)))},
\end{aligned}
\end{equation}
where $f(\cdot)$ is image encoder and $\text{s}(\cdot,\cdot)$ represents the similarity in the feature space, usually using cosine similarity.

By minimizing this loss on large-scale text-image pairs, the generated token embeddings will be well-aligned with the corresponding visual features.

\textbf{Stable Diffusion as Text-to-Image Decoder.} After obtaining the concept token embedding $\text{V}^*$, 
Text-to-Image Decoder generates an image $\text{x}_{gen}$ with an initial noise map $\varepsilon \sim \mathcal{N}(0,1)$ as:
\begin{equation}
\centering
\begin{aligned}
\text{x}_{gen} = \texttt{Image-Decoder}_{{x}_{\theta}}(\varepsilon,\text{V}^*). 
\end{aligned}
\end{equation}
Specifically, we use Stable Diffusion~\cite{rombach2022high} as the image decoder ${x}_{\theta}$. 
The concept token embedding $\text{V}^*$ is decoded into images by the cross-attention between textual and visual embeddings in the U-net decoder. The cross-attention layers project $\text{V}^*$ into keys $\mathbf{K}$ and values $\mathbf{V}$, while the queries $\mathbf{Q}$ are derived from the intermediate features of U-net. The attention maps are then calculated by $\mathbf{A}=\text{Softmax}(\frac{\mathbf{Q}\mathbf{K}^\intercal}{\sqrt{d}})$, where $d$ denotes the hidden state dimension. Finally, pixel features are
comprised of the values $\mathbf{V}$, weighted by the attention maps $\mathbf{A}$ as: $\mathbf{A}\cdot\mathbf{V}$. \cite{tewel2023key} find that the keys $\mathbf{K}$ control the compositional structure of the generated image, and the values $\mathbf{V}$ control the appearance of image components.

Stable Diffusion is trained using a squared error loss to denoise a variably-noised image or latent code $z_{t}$ as:
\begin{equation}
\centering
\begin{aligned}
\mathcal{L}_{rec} = \mathbb{E}_{\texttt{x}, \texttt{V}^{*}, \varepsilon, t} \left[ \| x_{\theta}(z_{t}, \text{V}^{*}) - \text{x} \|_2^2 \right],
\end{aligned}
\end{equation}
where $\text{x}$ is the ground-truth image, and $z_{t}=\sqrt{\alpha_{t}}\text{x}+\sqrt{1-\alpha_{t}}\varepsilon$ is the noisy input at time-step $t$ where $\alpha_{t}$ is related to a fixed variance schedule.
Such training objective can be simplified as a reconstruction loss. 

\begin{figure}
  \centering
  \includegraphics[scale=0.34]{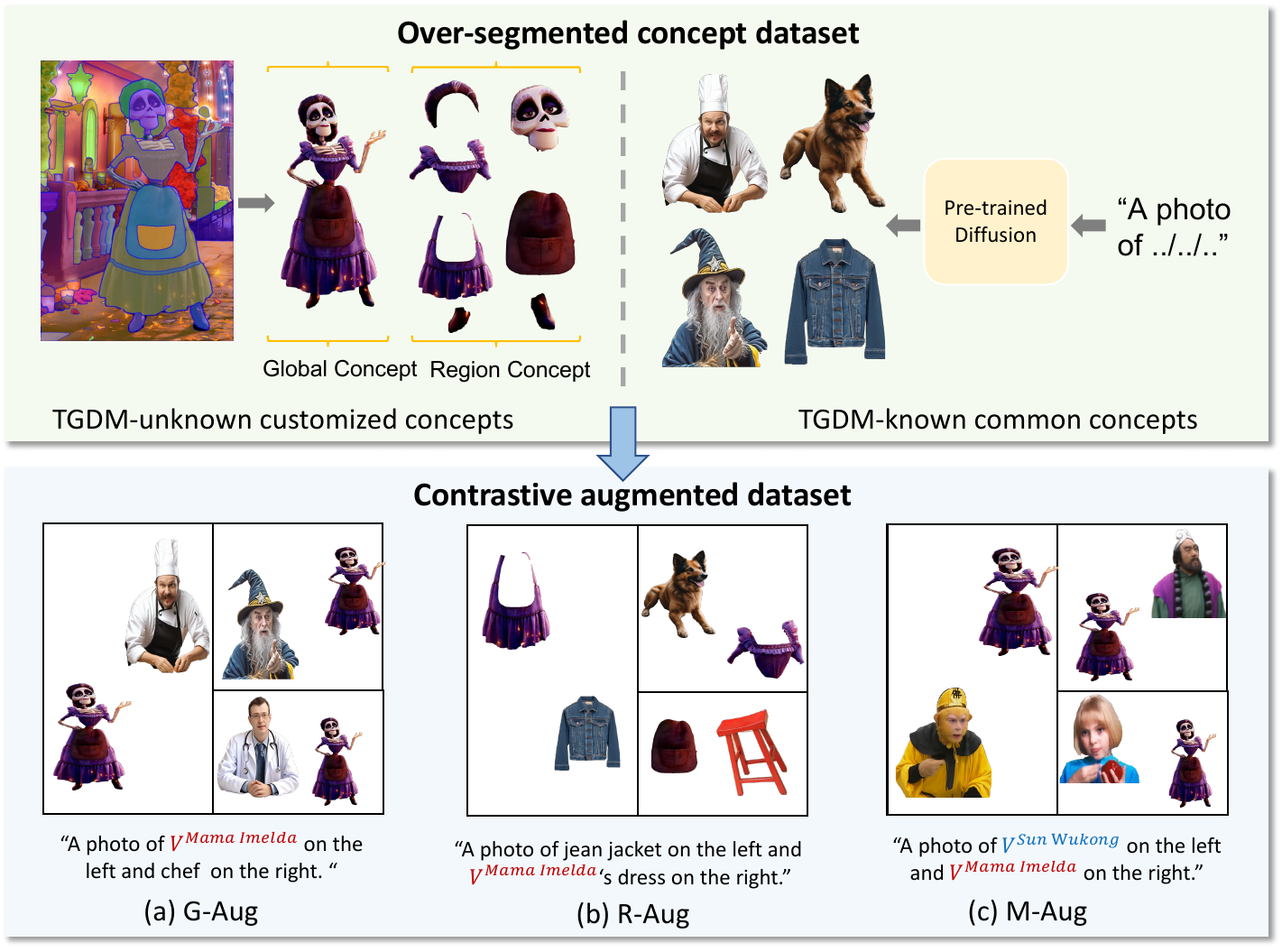}
\vspace{-1.5em}
\caption{Pipeline of training data curation. We mix the customized concepts and common concepts at instance-level and segmentation-level, to help decouple multi-concept token embeddings which can eliminate the confusion issues.}
\label{dataset}
\vspace{-1em}
\end{figure}

\section{Method: CLIF}

\begin{figure}
  \centering
  \includegraphics[scale=0.49]{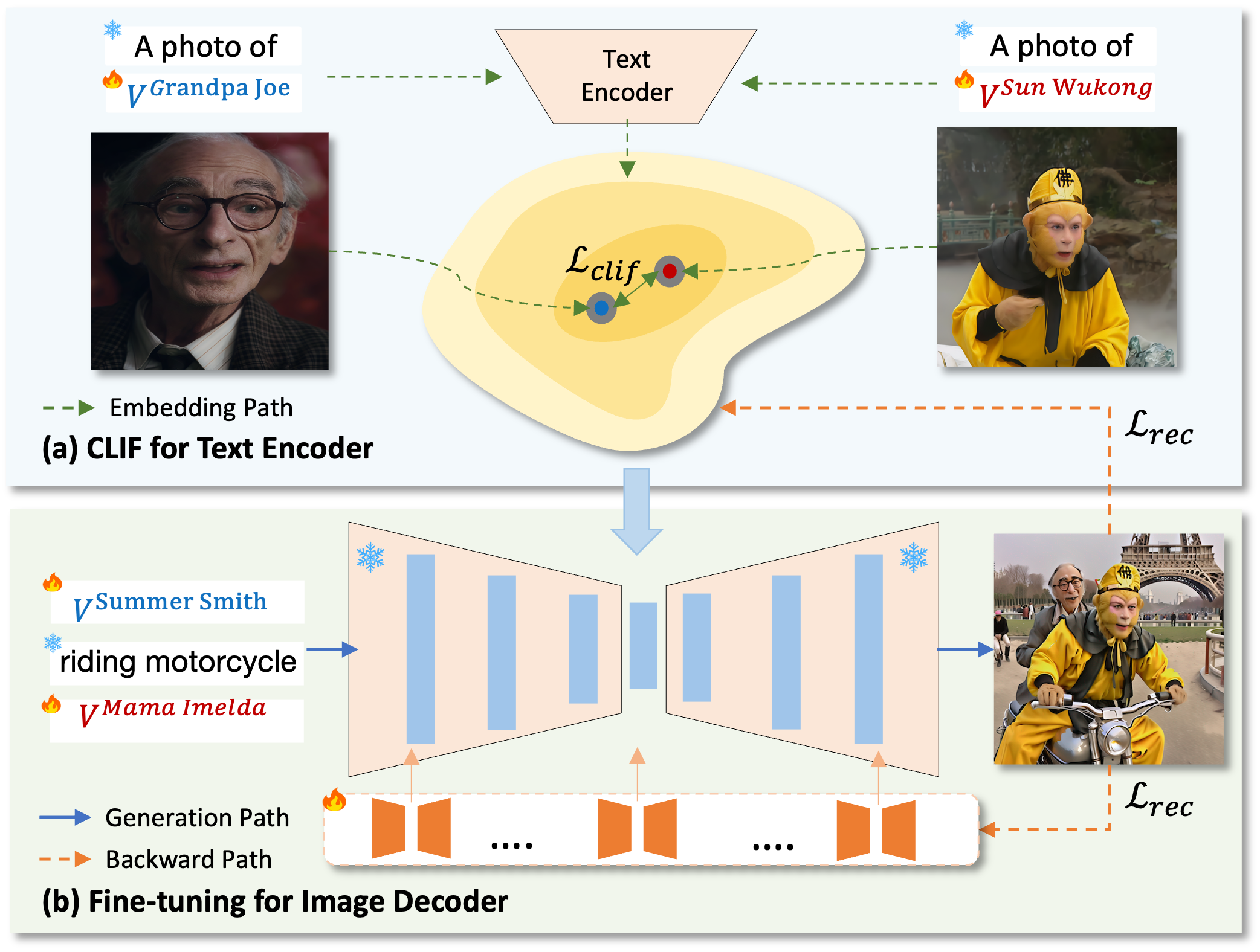}
\vspace{-0.5em}
\caption{Our two stage framework for multi-concept learning. We first fine-tune the text encoder to get contrastive concept embeddings, and then fine-tune the text-to-image decoder to synthesizing non-confusing images.}
\label{pipeline}
\vspace{-1.5em}
\end{figure}

We aim to compose multiple customized concepts in one image with complex interaction. To eliminate the confusion issues of baselines~\cite{gu2023mix,kumari2023multi} in Figure~\ref{main_1}, we re-visit the two stages of TGDM and then propose a two stage fine-tuning method as shown in Figure~\ref{pipeline}, to make the embeddings of different concepts more contrastive. To support an effective contrastive fine-tuning with very limited user-provided concept images, we design an over-segmented method 
with multi-granularity for training data curation to ensure the learned concept embeddings are separated globally and locally.

\subsection{Training Data Curation}
\label{4.1}
Upon revisiting the two stages of TGDM, 
it becomes apparent that the textual embeddings of newly added customized concepts (\textit{i.e.}, TGDM-unknown concepts) in existing customized generation methods are solely trained in the Text-To-Image Decoder stage and not in the Text Encoder stage.
They overlook that during image decoder training, the reconstruction loss $\mathcal{L}_{rec}$ aims to reconstruct visual features at the pixel level and should not be mistaken for the contrastive learning loss $\mathcal{L}_{clip}$, which decouples relationships between concepts. This results in the under-trained concept embedding and the projected $V$ being confusing, ultimately leading to confusion in generated multi-concept images.

Our idea is to fine-tune the customized concept embedding by contrastive learning similar to CLIP's to reduce its confusion. To this end, we propose a simple technique to construct a large number of image-text pairs for fine-tuning the text encoder and image decoder because the customized concepts are derived from the user-provided limited image data. Specifically, we decompose the confusion into three issues: 1) \textbf{Identity Preservation}, 2) \textbf{Attribute Binding}, and 3) \textbf{Concept Attendance}. To this end, as shown in Figure~\ref{dataset}, we construct three augmentation data: 1) Global Augmentation (\textbf{G-Aug}), 2) Region Augmentation (\textbf{R-Aug}), and 3) Mix Augmentation (\textbf{M-Aug}), to address the above issues respectively. Below, we describe in detail the motivation and process for constructing each type of data.

\textbf{G-Aug for Identity Preservation}
We draw the following two observations regarding existing approaches:
\textbf{1)} In Figure~\ref{confusion map}, we find that the degree of confusion and the Euclidean distance between embeddings show a correlation, which suggests that it is necessary to pull embeddings that use the same initialization (\textit{e.g.}, man) farther apart in the embedding space.
\textbf{2)} In Figure~\ref{main_1}, we observe that confusing concept embeddings will fail to preserve identity information, such as the erroneous fusion of \textit{TangSeng} and \textit{Hector Rivera}'s visual appearance. 

Based on the aforementioned observations, we attribute the identity loss to inter-concept confusion. To reduce it, we propose global augmentation.
Specifically, 1) we first segment the concept from the original images using SAM~\cite{kirillov2023segment}, to filter irrelevant contexts like the background; 2) then we utilize a pre-trained diffusion model to generate some general concepts such as policeman, dog, denim jacket, etc., as new contexts; 3) finally, we combine these segmented concepts with the general concepts to generate a large number of text-image pairs with different contexts.

The global augmentation is designed to fine-tune the text encoder. With the supervision contained in text paired with images, the concept token carries its visual features, excluding the visual features of other concepts, and inter-concept confusion is mitigated.

\textbf{R-Aug for Attribute Binding}.
In Figure~\ref{main_1}, we observe that the semantic components of a concept can also be confused, leading to attribute leakage, \textit{e.g.} \textit{Jerry Smith}'s top incorrectly uses the blue color of his pants.

Based on this observation, we attribute the attribute leakage to intra-concept confusion. To reduce it, we propose regional augmentation.
Specifically, 1) we first use GPT-4 to caption as much as possible the characterization in the concept such as hair, necklaces, hats, and so on; 2) then, we further segment the global concepts to get the regional concept, and label it with the results from GPT-4; 3) finally, we follow the global concepts' process to generate a large number of region-based text-image pairs.

The regional augmentation is designed to fine-tune text encoder and is complementary to global augmentation, working together to achieve a non-confusing text embedding.

\textbf{M-Aug for Concept Attendance}.
We make the following observations about missing concepts:
\textbf{1)} In Figure~\ref{main_1}, we observe that there are often dominant concepts, while other non-dominant concepts \textit{e.g.}, \textit{Miguel Ricveras} and \textit{Summer Smith}, are not always successfully generated in images, sometimes missing entirely.
\textbf{2)} We find the text-to-image decoder has defined the dominance in one of the concept embedding vectors beforehand, which we call dominant bias. For example, when given the prompt ``a photo of a cat and a pug", stable diffusion tends to generate two pugs due to the bias in the pre-train data.
The presence of dominant bias results in missing concepts in multiple concept generation where non-dominant concepts are often lost or produce redundant dominant concepts.

Based on the aforementioned observations, we attribute concept missing to the dominant bias. To reduce it, we propose mixed augmentation.
Specifically, 1) we first segment the concept from the original images similar to global augmentation; 2) then, we randomly scale and place the segmented concept with another one on either left or right side of the image and generate corresponding text prompts. 

The mixed augmentation is designed to fine-tune the text-to-image decoder. To present the model with correctly mixed image samples, the text-to-image decoder is enforced to synthesize multi-concepts equally.

\begin{figure*}
  \centering
  \includegraphics[scale=0.1]{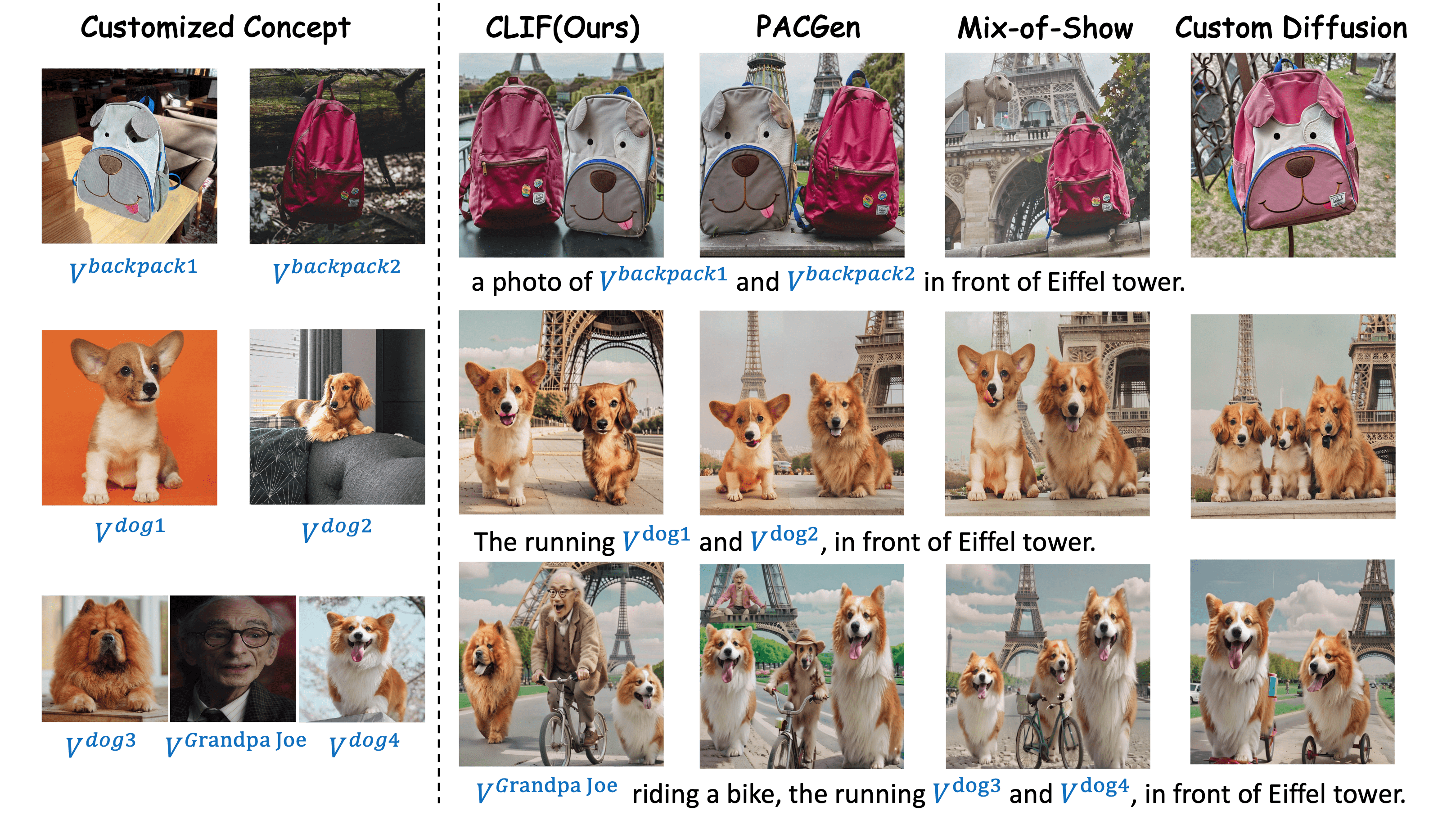}
\vspace{-1.5em}
\caption{Comparison of different methods on DreamBench.}
\label{dreambench}
\vspace{-1.5em}
\end{figure*}

\subsection{CLIF for Text Encoder}
\label{4.2}

We investigate embedding tuning~\cite{gal2022image,voynov2023p+} in concept customization. Given a text prompt containing the customized concept \textit{Jerry Smith} or \textit{Willy Wonka}, the supercategory embedding, \textit{e.g.}, ``man'', is used to initialize both concepts.

Our goal is to eliminate confusion in text embeddings by fine-tuning the customized concepts contrastively. The fine-tuning approach is similar to CLIP's training, optimizing a symmetric cross-entropy loss over these image-text similarity scores as follows:
\begin{equation}
\begin{aligned}
\mathcal{L}_{clif} = - \log \frac{\exp(\text{s}(\texttt{Text-Encoder}(p_{a}), f(q_{a})) )}{\sum_{q_{a}^-} \exp(\text{s}(\texttt{Text-Encoder}(p_{a}), f(q_{a}^-))) )},
\end{aligned}
\end{equation}
where $\text{s}(p_{a}, q_{a})$ is the similarity score between the augmented image $q_{a}$ and corresponding prompt $p_{a}$, while $\text{s}(p_{a}, q_{a}^{-})$ is the similarity score for negative pair in the batch.

\subsection{Fine-tuning for Text-to-Image Decoder}
\label{4.3}

After applying contrastive fine-tuning to the customized concepts in the text encoder, we have obtained decoupled concept embeddings $c$ and can generate customized images with well-maintained identity information. 

Our goal is to generate images that contain multi-customized concepts. However, existing weight fine-tuning techniques for the diffusion model are insufficient to achieve this goal by dominant bias. With the mixed concepts augmentation, we freeze the text encoder and jointly train multiple concepts with a shared LoRA $W_{\theta}$ in the U-net $\theta$. Since embeddings are already separated in the text encoder, during joint training the model can avoid concept conflict~\cite{gu2023mix}.

\section{Experimental Results}
\label{5}
\subsection{Experimental Setup}
\textbf{Task and Dataset.}
We aim to address the challenge of preventing the confusion of multiple customized concepts. To comprehensively verify the effectiveness of CLIF, we consider concepts as \textit{characters} comprising a range of visual elements (\textit{e.g.}, ``face'', ``hat'', and ``clothes'') that need to be preserved. We curate a dataset consisting of $18$ representative characters, including $9$ real-world, $4$ 3D-animated, and $5$ 2D-animated. Each of them possesses unique visual appearances that must be preserved in the customized generation. In our experiment, we will demonstrate the ability of CLIF to generate imaginative compositions of these characters with complex interactions involving spatial clutter (\textit{e.g.}, ``snuggling'' and ``riding motorcycle'').

\textbf{Baselines.}
We compare CLIF against state-of-the-art baselines: Text-Inversion~\cite{gal2022image}, Custom Diffusion~\cite{kumari2023multi}, Dreambooth~\cite{ruiz2023dreambooth}, and Mix-of-Show~\cite{gu2023mix}. 
Moreover, to demonstrate the generalizability of the proposed strategy in CLIF, we integrate it with Text-Inversion and Custom Diffusion. Note that for fair comparison, all methods do not incorporate additional spatial constraints as in~\cite{zhang2023adding}. 

\begin{figure*}
  \centering
  \includegraphics[scale=0.47]{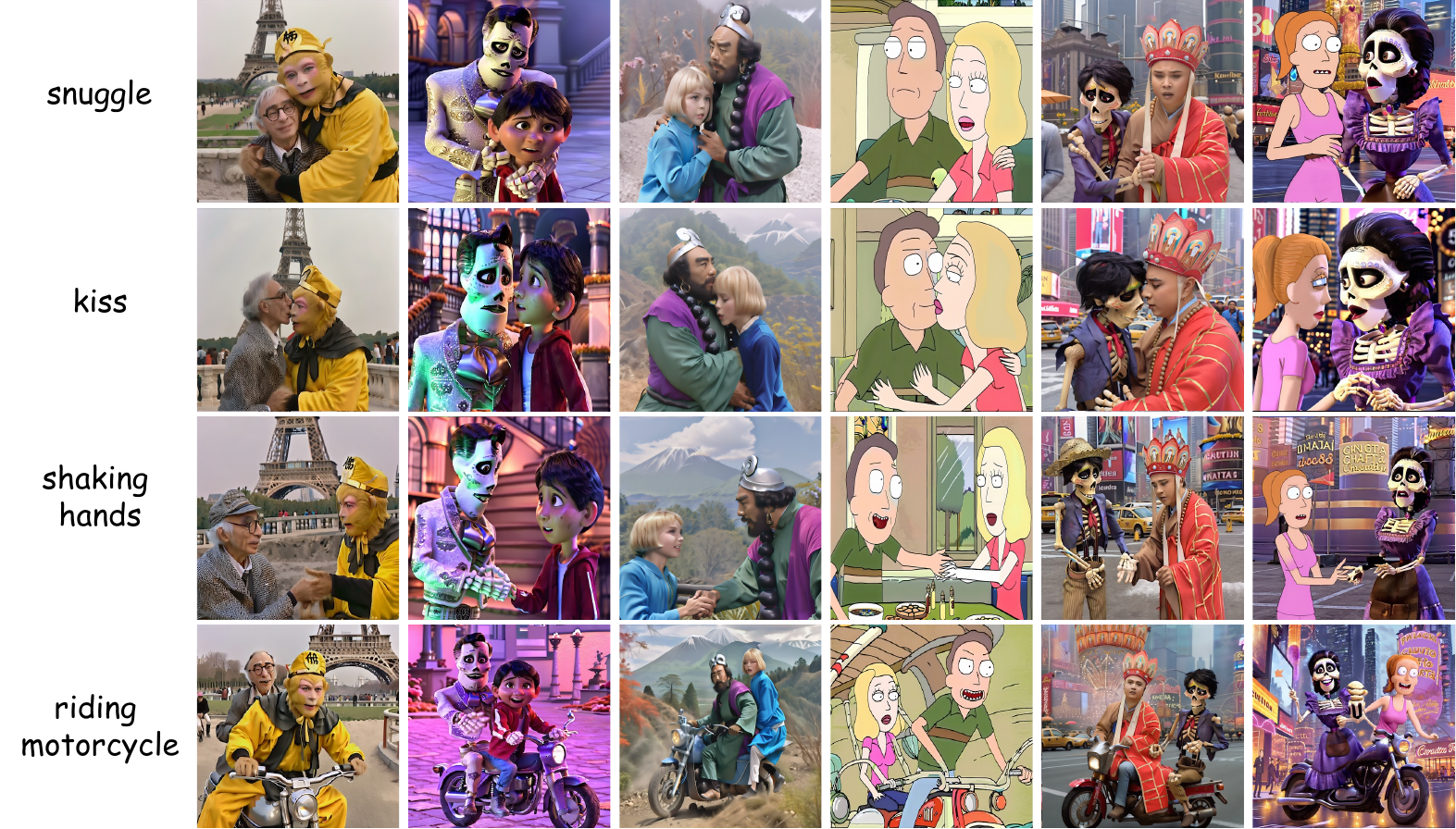}
\vspace{-1em}
\caption{Results for multi-concept customized generation using CLIF. Our approach is able to generate non-confusion images containing multiple characters with complex interactions, without the need for additional spatial constraints (\textit{e.g.}, layout, mask, sketch).}
\label{multi}
\vspace{-1.5em}
\end{figure*}

\subsection{Qualitative Comparison}
We compare CLIF with Mix-of-Show, Custom Diffusion (Custom for short), and Text-Inversion (TI for short) for multi-concept customized generation in Figure \ref{main_1}.
The baseline methods suffer from, identity loss (highlighted in red box), attribute leaking (highlighted in blue box), or concept missing (highlighted in green box).
TI and Custom implicitly delegate the task of disentangling multi-concept token embeddings to the Text-to-Image decoder. However, this paradigm is limited to the reconstruction loss which only encodes the concept's visual features into the embeddings without contrasting it with other token embeddings.
On the other hand, Mix-of-Show uses gradient fusion to merge multiple separately fine-tuned concepts, which aims to preserve the single concept identity in the fused model rather than decoupling each other.

To demonstrate the generalization of our method, we conduct experiments on Dreambench. We incorporate an additional baseline PACGen\cite{li2023generate} which is a multi-subject driven generation method in Dreambench. The experimental results in Figure \ref{dreambench} demonstrate that our approach generalizes well to Dreambench.

Based on the results of CLIF for multi-concept customization in Figure~\ref{multi}, we can find that:
1) Benefiting from contrastive fine-tuning, CLIF can accurately generate each character even when all the customized concepts belong to the same superclass (\textit{i.e.}, ``human''). This indicates CLIF's ability to differentiate between similar concepts within the same category; and 
2) In the case of multi-concept generation with complex interactions (such as ``snuggling'' and ``kissing''), previous approaches exhibit more serious confusion problems. 
Due to the spatial clutter between concepts, these approaches tend to draw visual features of multiple concepts in duplicate areas, which results in the wrong combination or loss of concepts. 
CLIF can handle complex interactions natively through contrastive fine-tuned text embedding, without relying on additional spatial constraints, making it a more cost-effective solution.
3) These results demonstrate the effectiveness of directly contrasting the textual embeddings of customized concepts in the first stage, which reduces the confusion significantly. Generating more than 2 objects can be achieved by simply extending our dataset. However, it has some limitations which are discussed in Appendix.

\subsection{Quantitative Comparison}
Following Custom Diffusion~\cite{kumari2023multi}, we utilize the text/image encoder of CLIP to assess text alignment and image alignment.  
A detailed evaluation setting is provided in Appendix.

\begin{table}[t]
\centering
\caption{Text-alignment and image-alignment vary between single-concept and multi-concept generation scenarios.}
\scalebox{0.68}{
\begin{tabular}{lcccc}
\toprule
\multirow{2}{*}{Methods} & \multicolumn{2}{c}{Text Alignment} & \multicolumn{2}{c}{Image Alignment} \\ \cline{2-5}
                         & Single & Multi  & Single & Multi       \\
 \hline
TI             &$0.604$(\textcolor{blue}{-$2.6$\%})                 &$0.507$(\textcolor{blue}{-$9.2$\%})                 &$0.726$(\textcolor{blue}{-$6.1$\%})                  &$0.708$(\textcolor{blue}{-$5.3$\%})                        \\
DreamBooth     &$0.617$(\textcolor{blue}{-$1.3$\%})                 &$0.523$(\textcolor{blue}{-$7.6$\%})                 &$0.754$(\textcolor{blue}{-$3.3$\%})                  &$0.711$(\textcolor{blue}{-$5.0$\%})                          \\
Custom         &$0.622$(\textcolor{blue}{-$0.8$\%})                 &$0.511$(\textcolor{blue}{-$8.8$\%})                 &$0.749$(\textcolor{blue}{-$3.8$\%})                  &$0.715$(\textcolor{blue}{-$4.6$\%})                          \\
Mix-of-Show    &$0.629$(\textcolor{blue}{-$0.1$\%})                 &$0.526$(\textcolor{blue}{-$7.3$\%})                 &$0.757$(\textcolor{blue}{-$3.0$\%})                  &$0.713$(\textcolor{blue}{-$4.8$\%})                         \\ \hline
TI+CLIF        &$0.631$(\textcolor{red}{+$0.1$\%})                 &$0.528$(\textcolor{blue}{-$7.1$\%})                 &$0.751$(\textcolor{blue}{-$3.6$\%})                  &$0.726$(\textcolor{blue}{-$3.5$\%})                          \\
Custom+CLIF    &$0.657$(\textcolor{red}{+$2.7$\%})                 &$0.535$(\textcolor{blue}{-$6.4$\%})                 &$0.774$(\textcolor{blue}{-$1.3$\%})                  &$0.730$(\textcolor{blue}{-$3.1$\%})                          \\ \hline
CLIF (ours)     &$0.630$            &$0.599$            &$0.787$             &$0.761$                     \\
\bottomrule
\end{tabular}
}
\vspace{-1em}
\label{compare}
\end{table}

\begin{table}[]
\centering
\setlength{\belowcaptionskip}{-0.6cm}
\caption{Quantitative ablation study in single-concept and multi-concept generation scenarios.}
\scalebox{0.72}{
\begin{tabular}{lcccccc}
\toprule
\multirow{2}{*}{Methods} & \multicolumn{2}{c}{Text Alignment} & \multicolumn{2}{c}{Image Alignment}  \\ \cline{2-5} 
                         & Single & Multi &Single  & Multi     \\
                         \hline                      
CLIF     &$0.630$                  &$0.599$                 &$0.787$                 &$0.761$                                   \\
\hline
w/o G-Aug         &$0.604$(\textcolor{blue}{-$2.6$\%})                  &$0.566$(\textcolor{blue}{-$3.3$\%})                 &$0.751$(\textcolor{blue}{-$3.6$\%})                  &$0.729$(\textcolor{blue}{-$3.2$\%})                                     \\
w/o R-Aug         &$0.627$(\textcolor{blue}{-$0.3$\%})                  &$0.591$(\textcolor{blue}{-$0.8$\%})                 &$0.774$(\textcolor{blue}{-$1.3$\%})                  &$0.753$(\textcolor{blue}{-$0.8$\%})                                    \\
w/o M-Aug     &$0.612$(\textcolor{blue}{-$1.8$\%})                  &$0.579$(\textcolor{blue}{-$2.0$\%})                 &$0.768$(\textcolor{blue}{-$1.9$\%})                 &$0.740$(\textcolor{blue}{-$2.1$\%})                    \\
\bottomrule
\end{tabular}
}
\vspace{-1em}
\label{table_ablation}
\end{table}

Based on the results presented in Table \ref{compare}, we can find that:
1) For single-concept, compared with TI which encodes all concept details within the text embedding, CLIF and other baseline methods benefit from tuning the diffusion weight and exhibit superior image alignment;
2) For single-concept, CLIF exhibits superior image alignment compared to baseline methods. This is attributed to the contrastive fine-tuning of textual embeddings in CLIF. Contrastive fine-tuning not only helps mitigate confusion but also aligns the image and the concept, enabling the token embeddings of concept names to capture more detailed visual features.
Furthermore, CLIF maintains comparable text alignment, indicating that our approach can enhance high identity preservation without compromising composability;
3) For multi-concept, the superior performance of CLIF across all metrics highlights its ability to effectively capture concept characteristics and preserve concept distinct identity in multi-concept compositions; and 
4) The integration of CLIF with TI and Custom shows a remarkable improvement, indicating that contrastive fine-tuning of the text embedding is indeed effective and generalizable.

\begin{figure}
  \centering
  \includegraphics[scale=0.88]{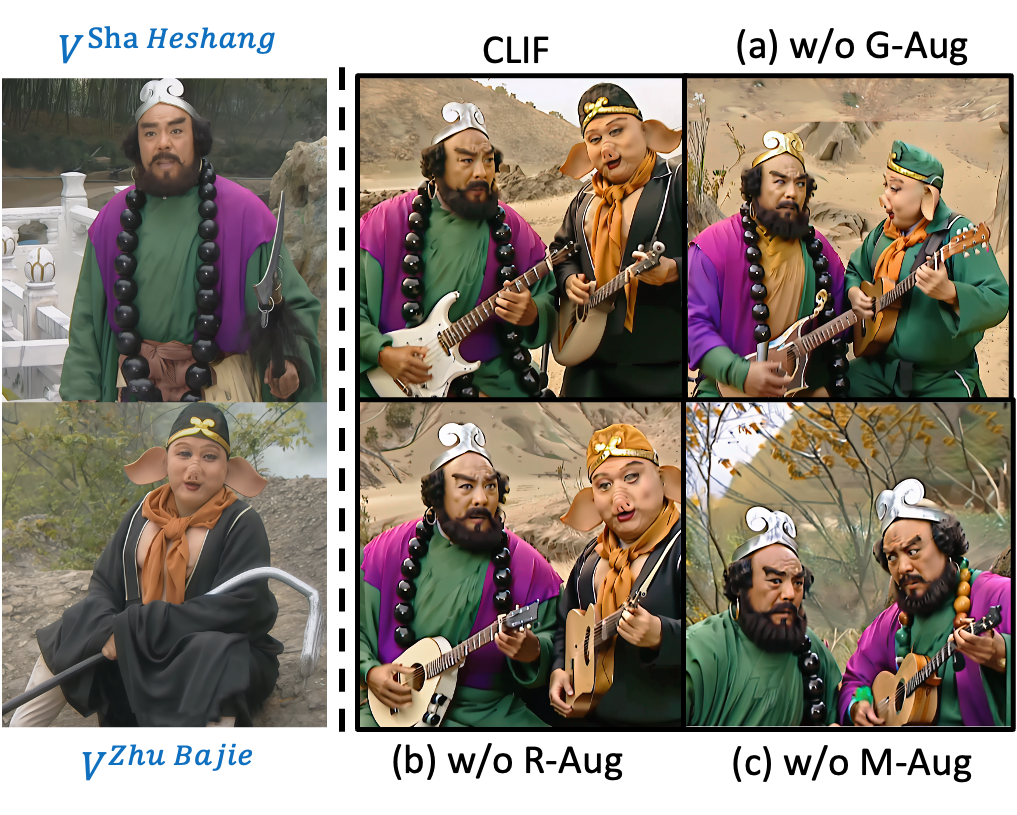}
\vspace{-1em}
\caption{Visualization of ablation results for augmented data type.}
\label{ablation}
\vspace{-1em}
\end{figure}

\subsection{Ablation Study}

As mentioned in Section~\ref{4.1}, three critical capabilities must be addressed to mitigate the confusion problem in customized generation, namely, \textit{identity preservation}, \textit{attribute binding}, and \textit{concept attendance}. The following ablation experiments will demonstrate how CLIF is specifically designed to enhance these capabilities.

\textbf{Effectiveness of Global Augmentation (G-Aug).} In Figure \ref{ablation}, it can be observed that without global augmentation, the generation results suffer from identity departure (\textit{e.g.}, \textit{Zhu Bajie} incorrectly uses \textit{Sha Heshang}'s coat color). In contrast, our CLIF successfully mitigates this issue by pushing the textual embeddings of the two concepts farther apart through contrastive supervision, which prevents the Text-to-Image Decoder from confusing the characters' visual appearances during generation. 
This interpretation is also supported by the quantitative results in Table \ref{table_ablation}, which indicate that the image alignment in multi-concept scenario decreases from $0.761$ to $0.729$ (-$3.2$\%).

\textbf{Effectiveness of Region Augmentation (R-Aug).}
Even when multiple concepts are decoupled from each other, the generated concepts may still struggle with attribute leakage ( \textit{e.g.}, \textit{Zhu Bajie}'s hat incorrectly using the color of his tie). Therefore, we use region augmentation to fine-tune each component of the embedding by binding each sub-region of the concept to the token embedding, which not only decouples the confusion of similar components between concepts but also decouples the confusion within concepts. According to the results in Table \ref{table_ablation},  region augmentation improves attribute binding during multi-concept generation, boosting both single and multi-concept image alignment.

\textbf{Effectiveness of Mix Augmentation (M-Aug).}
A common problem in multi-concept generation is concept missing. As shown in Figure \ref{ablation}, two instances of \textit{Sha Heshang} were repeatedly generated while \textit{Zhu Bajie} was missing. We attribute this issue to dominant bias and address it through mix augmentation. 
To highlight the impact of mix augmentation on addressing dominant bias, we propose an intuitive metric \textit{attendance}, as described in Appendix. The results are shown in Figure~\ref{bais}, where the attendance score for all $18$ customized concepts is significantly improved, demonstrating the effectiveness of mix augmentation.

\begin{figure}
  \centering
  \includegraphics[scale=0.445]{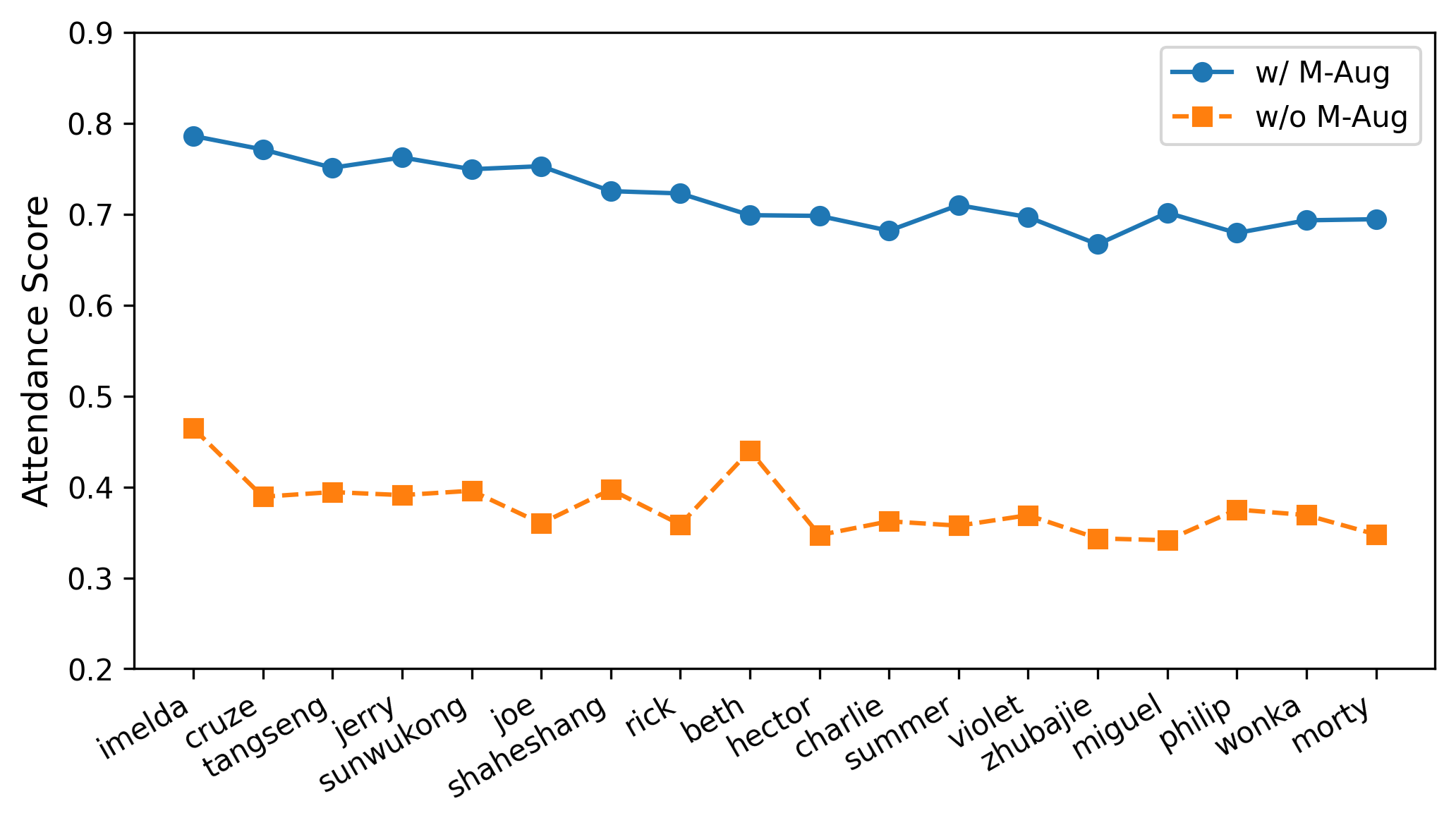}
\vspace{-1em}
\caption{Effects of using M-Aug for concept missing.}
\label{bais}
\vspace{-0.5em}
\end{figure}

\textbf{Cross-Attention in Diffusion.}
\begin{figure}
  \centering
  \includegraphics[scale=0.23]{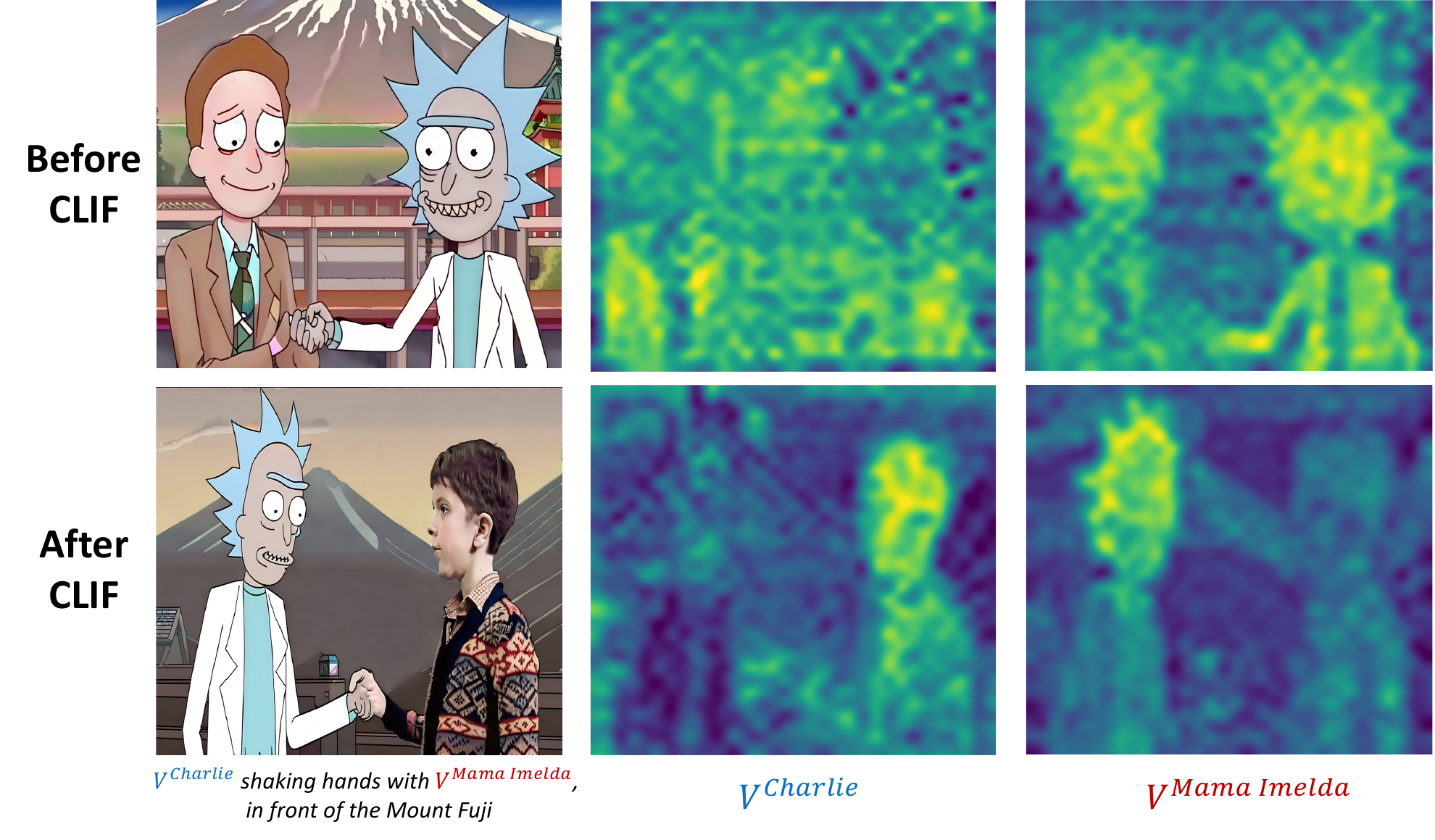}
\vspace{-0.5em}
\caption{Visualization of attention map for concept embeddings.}
\label{augviews}
\vspace{-1em}
\end{figure}
We visualize concept tokens' cross-attention maps before and after CLIF in Figure~\ref{augviews}. The results indicate that contrastive fine-tuning effectively decouples the token embeddings of multiple customized concepts, eliminates confusion in cross-attention maps, and generates high-quality images.

\section{Conclusions}
Our CLIF approach marks a significant advancement in the customized generative models. By fine-tuning both the text encoder and text-to-image decoder stages, CLIF successfully addresses the persistent challenge of concept confusion, particularly in multi-concept generation scenarios. This technique preserves the integrity of each concept, ensuring that each retains its unique identity even amidst complex and cluttered interactions.  Our extensive experiments and ablation studies underscore the efficacy of CLIF, establishing it as a powerful and versatile tool for customized concept generation. The improvements observed in both single and multi-concept customizations indicate the broad applicability and potential of our method in various creative and practical applications. 

\section{Impact Statements}
\textbf{Ethical Impacts} This study does not raise any ethical concerns. The research does not involve subjective assessments or the use of private data. Only publicly available datasets are utilized for experimentation.

\textbf{Expected Societal Implications}
We aim to address the confusion in customized concepts and facilitate the generation of higher-quality customized multi-concept images. A primary ethical concern is the potential misuse of this technology, notably in creating deepfakes, which can result in misinformation, privacy violations, and other harmful outcomes. To mitigate these risks, the establishment of robust ethical guidelines and continuous monitoring is essential. 

The concern raised here is a common one, not just for our method but across various multi-concept customization techniques. A viable strategy to lessen these risks might be the implementation of tactics akin to those used in anti-dreambooth\cite{van2023anti}. This approach involves adding minor noise disturbances to the shared images, thereby hindering the customization process. Furthermore, embedding invisible watermarks in the generated images can serve as a deterrent against misuse and ensure that they are not used without due acknowledgment.

\nocite{langley00}

\bibliography{example_paper}

\begin{thebibliography}{28}
\providecommand{\natexlab}[1]{#1}
\providecommand{\url}[1]{\texttt{#1}}
\expandafter\ifx\csname urlstyle\endcsname\relax
  \providecommand{\doi}[1]{doi: #1}\else
  \providecommand{\doi}{doi: \begingroup \urlstyle{rm}\Url}\fi

\bibitem[Alaluf et~al.(2023)Alaluf, Richardson, Metzer, and Cohen-Or]{alaluf2023neural}
Alaluf, Y., Richardson, E., Metzer, G., and Cohen-Or, D.
\newblock A neural space-time representation for text-to-image personalization.
\newblock \emph{ACM Transactions on Graphics (TOG)}, 42\penalty0 (6):\penalty0 1--10, 2023.

\bibitem[Balaji et~al.(2022)Balaji, Nah, Huang, Vahdat, Song, Kreis, Aittala, Aila, Laine, Catanzaro, et~al.]{balaji2022ediffi}
Balaji, Y., Nah, S., Huang, X., Vahdat, A., Song, J., Kreis, K., Aittala, M., Aila, T., Laine, S., Catanzaro, B., et~al.
\newblock ediffi: Text-to-image diffusion models with an ensemble of expert denoisers.
\newblock \emph{arXiv preprint arXiv:2211.01324}, 2022.

\bibitem[Chefer et~al.(2023)Chefer, Alaluf, Vinker, Wolf, and Cohen-Or]{chefer2023attend}
Chefer, H., Alaluf, Y., Vinker, Y., Wolf, L., and Cohen-Or, D.
\newblock Attend-and-excite: Attention-based semantic guidance for text-to-image diffusion models.
\newblock \emph{ACM Transactions on Graphics (TOG)}, 42\penalty0 (4):\penalty0 1--10, 2023.

\bibitem[Feng et~al.(2022)Feng, He, Fu, Jampani, Akula, Narayana, Basu, Wang, and Wang]{feng2022training}
Feng, W., He, X., Fu, T.-J., Jampani, V., Akula, A., Narayana, P., Basu, S., Wang, X.~E., and Wang, W.~Y.
\newblock Training-free structured diffusion guidance for compositional text-to-image synthesis.
\newblock \emph{arXiv preprint arXiv:2212.05032}, 2022.

\bibitem[Gal et~al.(2022)Gal, Alaluf, Atzmon, Patashnik, Bermano, Chechik, and Cohen-Or]{gal2022image}
Gal, R., Alaluf, Y., Atzmon, Y., Patashnik, O., Bermano, A.~H., Chechik, G., and Cohen-Or, D.
\newblock An image is worth one word: Personalizing text-to-image generation using textual inversion.
\newblock \emph{arXiv preprint arXiv:2208.01618}, 2022.

\bibitem[Gu et~al.(2023)Gu, Wang, Wu, Shi, Chen, Fan, Xiao, Zhao, Chang, Wu, et~al.]{gu2023mix}
Gu, Y., Wang, X., Wu, J.~Z., Shi, Y., Chen, Y., Fan, Z., Xiao, W., Zhao, R., Chang, S., Wu, W., et~al.
\newblock Mix-of-show: Decentralized low-rank adaptation for multi-concept customization of diffusion models.
\newblock \emph{arXiv preprint arXiv:2305.18292}, 2023.

\bibitem[Hertz et~al.(2022)Hertz, Mokady, Tenenbaum, Aberman, Pritch, and Cohen-Or]{hertz2022prompt}
Hertz, A., Mokady, R., Tenenbaum, J., Aberman, K., Pritch, Y., and Cohen-Or, D.
\newblock Prompt-to-prompt image editing with cross attention control.
\newblock \emph{arXiv preprint arXiv:2208.01626}, 2022.

\bibitem[Hu et~al.(2021)Hu, Shen, Wallis, Allen-Zhu, Li, Wang, Wang, and Chen]{hu2021lora}
Hu, E.~J., Shen, Y., Wallis, P., Allen-Zhu, Z., Li, Y., Wang, S., Wang, L., and Chen, W.
\newblock Lora: Low-rank adaptation of large language models.
\newblock \emph{arXiv preprint arXiv:2106.09685}, 2021.

\bibitem[Huang et~al.(2023)Huang, Chen, Liu, Shen, Zhao, and Zhou]{huang2023composer}
Huang, L., Chen, D., Liu, Y., Shen, Y., Zhao, D., and Zhou, J.
\newblock Composer: Creative and controllable image synthesis with composable conditions.
\newblock \emph{arXiv preprint arXiv:2302.09778}, 2023.

\bibitem[Kirillov et~al.(2023)Kirillov, Mintun, Ravi, Mao, Rolland, Gustafson, Xiao, Whitehead, Berg, Lo, et~al.]{kirillov2023segment}
Kirillov, A., Mintun, E., Ravi, N., Mao, H., Rolland, C., Gustafson, L., Xiao, T., Whitehead, S., Berg, A.~C., Lo, W.-Y., et~al.
\newblock Segment anything.
\newblock \emph{arXiv preprint arXiv:2304.02643}, 2023.

\bibitem[Kumari et~al.(2023)Kumari, Zhang, Zhang, Shechtman, and Zhu]{kumari2023multi}
Kumari, N., Zhang, B., Zhang, R., Shechtman, E., and Zhu, J.-Y.
\newblock Multi-concept customization of text-to-image diffusion.
\newblock In \emph{Proceedings of the IEEE/CVF Conference on Computer Vision and Pattern Recognition}, pp.\  1931--1941, 2023.

\bibitem[Li et~al.(2023{\natexlab{a}})Li, Keuper, Zhang, and Khoreva]{li2023divide}
Li, Y., Keuper, M., Zhang, D., and Khoreva, A.
\newblock Divide \& bind your attention for improved generative semantic nursing.
\newblock \emph{arXiv preprint arXiv:2307.10864}, 2023{\natexlab{a}}.

\bibitem[Li et~al.(2023{\natexlab{b}})Li, Liu, Wen, and Lee]{li2023generate}
Li, Y., Liu, H., Wen, Y., and Lee, Y.~J.
\newblock Generate anything anywhere in any scene.
\newblock \emph{arXiv preprint arXiv:2306.17154}, 2023{\natexlab{b}}.

\bibitem[Liu et~al.(2022)Liu, Li, Du, Torralba, and Tenenbaum]{liu2022compositional}
Liu, N., Li, S., Du, Y., Torralba, A., and Tenenbaum, J.~B.
\newblock Compositional visual generation with composable diffusion models.
\newblock In \emph{European Conference on Computer Vision}, pp.\  423--439. Springer, 2022.

\bibitem[Liu et~al.(2023)Liu, Zhang, Shen, Zheng, Zhu, Feng, Liu, Zhao, Zhou, and Cao]{liu2023cones}
Liu, Z., Zhang, Y., Shen, Y., Zheng, K., Zhu, K., Feng, R., Liu, Y., Zhao, D., Zhou, J., and Cao, Y.
\newblock Cones 2: Customizable image synthesis with multiple subjects.
\newblock \emph{arXiv preprint arXiv:2305.19327}, 2023.

\bibitem[Mou et~al.(2023)Mou, Wang, Xie, Zhang, Qi, Shan, and Qie]{mou2023t2i}
Mou, C., Wang, X., Xie, L., Zhang, J., Qi, Z., Shan, Y., and Qie, X.
\newblock T2i-adapter: Learning adapters to dig out more controllable ability for text-to-image diffusion models.
\newblock \emph{arXiv preprint arXiv:2302.08453}, 2023.

\bibitem[Patashnik et~al.(2023)Patashnik, Garibi, Azuri, Averbuch-Elor, and Cohen-Or]{patashnik2023localizing}
Patashnik, O., Garibi, D., Azuri, I., Averbuch-Elor, H., and Cohen-Or, D.
\newblock Localizing object-level shape variations with text-to-image diffusion models.
\newblock \emph{arXiv preprint arXiv:2303.11306}, 2023.

\bibitem[Po et~al.(2023)Po, Yang, Aberman, and Wetzstein]{po2023orthogonal}
Po, R., Yang, G., Aberman, K., and Wetzstein, G.
\newblock Orthogonal adaptation for modular customization of diffusion models.
\newblock \emph{arXiv preprint arXiv:2312.02432}, 2023.

\bibitem[Radford et~al.(2021)Radford, Kim, Hallacy, Ramesh, Goh, Agarwal, Sastry, Askell, Mishkin, Clark, et~al.]{radford2021learning}
Radford, A., Kim, J.~W., Hallacy, C., Ramesh, A., Goh, G., Agarwal, S., Sastry, G., Askell, A., Mishkin, P., Clark, J., et~al.
\newblock Learning transferable visual models from natural language supervision.
\newblock In \emph{International conference on machine learning}, pp.\  8748--8763. PMLR, 2021.

\bibitem[Rombach et~al.(2022)Rombach, Blattmann, Lorenz, Esser, and Ommer]{rombach2022high}
Rombach, R., Blattmann, A., Lorenz, D., Esser, P., and Ommer, B.
\newblock High-resolution image synthesis with latent diffusion models.
\newblock In \emph{Proceedings of the IEEE/CVF conference on computer vision and pattern recognition}, pp.\  10684--10695, 2022.

\bibitem[Ruiz et~al.(2023)Ruiz, Li, Jampani, Pritch, Rubinstein, and Aberman]{ruiz2023dreambooth}
Ruiz, N., Li, Y., Jampani, V., Pritch, Y., Rubinstein, M., and Aberman, K.
\newblock Dreambooth: Fine tuning text-to-image diffusion models for subject-driven generation.
\newblock In \emph{Proceedings of the IEEE/CVF Conference on Computer Vision and Pattern Recognition}, pp.\  22500--22510, 2023.

\bibitem[Ryu(2023)]{ryu2023low}
Ryu, S.
\newblock Low-rank adaptation for fast text-to-image diffusion fine-tuning, 2023.

\bibitem[Schuhmann et~al.(2021)Schuhmann, Vencu, Beaumont, Kaczmarczyk, Mullis, Katta, Coombes, Jitsev, and Komatsuzaki]{schuhmann2021laion}
Schuhmann, C., Vencu, R., Beaumont, R., Kaczmarczyk, R., Mullis, C., Katta, A., Coombes, T., Jitsev, J., and Komatsuzaki, A.
\newblock Laion-400m: Open dataset of clip-filtered 400 million image-text pairs.
\newblock \emph{arXiv preprint arXiv:2111.02114}, 2021.

\bibitem[Tewel et~al.(2023)Tewel, Gal, Chechik, and Atzmon]{tewel2023key}
Tewel, Y., Gal, R., Chechik, G., and Atzmon, Y.
\newblock Key-locked rank one editing for text-to-image personalization.
\newblock In \emph{ACM SIGGRAPH 2023 Conference Proceedings}, pp.\  1--11, 2023.

\bibitem[Van~Le et~al.(2023)Van~Le, Phung, Nguyen, Dao, Tran, and Tran]{van2023anti}
Van~Le, T., Phung, H., Nguyen, T.~H., Dao, Q., Tran, N.~N., and Tran, A.
\newblock Anti-dreambooth: Protecting users from personalized text-to-image synthesis.
\newblock In \emph{Proceedings of the IEEE/CVF International Conference on Computer Vision}, pp.\  2116--2127, 2023.

\bibitem[Voynov et~al.(2023)Voynov, Chu, Cohen-Or, and Aberman]{voynov2023p+}
Voynov, A., Chu, Q., Cohen-Or, D., and Aberman, K.
\newblock $ p+ $: Extended textual conditioning in text-to-image generation.
\newblock \emph{arXiv preprint arXiv:2303.09522}, 2023.

\bibitem[Yuan et~al.(2023)Yuan, Cun, Zhang, Li, Qi, Wang, Shan, and Zheng]{yuan2023inserting}
Yuan, G., Cun, X., Zhang, Y., Li, M., Qi, C., Wang, X., Shan, Y., and Zheng, H.
\newblock Inserting anybody in diffusion models via celeb basis.
\newblock \emph{arXiv preprint arXiv:2306.00926}, 2023.

\bibitem[Zhang et~al.(2023)Zhang, Rao, and Agrawala]{zhang2023adding}
Zhang, L., Rao, A., and Agrawala, M.
\newblock Adding conditional control to text-to-image diffusion models.
\newblock In \emph{Proceedings of the IEEE/CVF International Conference on Computer Vision}, pp.\  3836--3847, 2023.

\end{thebibliography}
\bibliographystyle{icml2024}

\end{document}